\documentclass[journal]{vgtc}                
\ifpdf
  \pdfoutput=1\relax                   
  \usepackage{graphicx}                
  \DeclareGraphicsExtensions{.pdf,.png,.jpg,.jpeg} 
\else
  \usepackage{graphicx}                
  \DeclareGraphicsExtensions{.eps}     
\fi%

\graphicspath{{figures/}{pictures/}{images/}{./}} 
\usepackage{amsmath}
\usepackage{microtype}                 
\PassOptionsToPackage{warn}{textcomp}  
\usepackage{textcomp}                  
\usepackage{mathptmx}                  
\usepackage{times}                     
\usepackage{cite}                      

\usepackage{cite}
\usepackage{mathptmx}
\usepackage{graphicx}
\usepackage{times}
\usepackage{multirow}
\usepackage{colortbl} 
\usepackage{tabularx}
\usepackage{enumitem}
\usepackage[switch]{lineno}
\usepackage{amssymb}
\usepackage{tabu}                      
\usepackage{booktabs}                  
\usepackage{lipsum}                    
\usepackage{mwe}

\newcommand{\revise}[1]{{\color{black}{#1}}}
\newcommand{\eg}{\emph{e.g.}}
\newcommand{\ie}{\emph{i.e.}}

\definecolor{red_color}{RGB}{194,40,45}
\newcommand{\redColor}[1]{\textcolor{red_color}{#1}}

\definecolor{green_color}{RGB}{0,146,70}
\newcommand{\greenColor}[1]{\textcolor{green_color}{#1}}

\onlineid{1193}

\vgtccategory{Research}
\vgtcpapertype{please specify}

\title{Advancing Multimodal Large Language Models in Chart Question Answering with Visualization-Referenced Instruction Tuning}

\author{Xingchen Zeng, Haichuan Lin, Yilin Ye, Wei Zeng, \textit{Member, IEEE}}
\authorfooter{
\item
Xingchen Zeng, Haichuan Lin, Yilin Ye, Wei Zeng are with the Hong Kong University of Science and Technology (Guangzhou), Guangzhou, China. Yilin Ye and Wei Zeng are also with the Hong Kong University of Science and Technology, Hong Kong SAR, China. E-mail: \{xzeng159@connect., hlin386@connect., yyebd@connect., weizeng@\}hkust-gz.edu.cn

\item  Wei Zeng is the corresponding author.
 }

\shortauthortitle{Biv \MakeLowercase{\textit{et al.}}: Global Illumination for Fun and Profit}

\abstract{
Emerging multimodal large language models (MLLMs) exhibit great potential for chart question answering (CQA). 
Recent efforts primarily focus on scaling up training datasets (\ie, charts, data tables, and question-answer (QA) pairs) through data collection and synthesis. 
However, our empirical study on existing MLLMs and CQA datasets reveals notable gaps. 
First, current data collection and synthesis focus on data volume and lack consideration of fine-grained visual encodings and QA tasks, resulting in unbalanced data distribution divergent from practical CQA scenarios. 
Second, existing work follows the training recipe of the base MLLMs initially designed for natural images, under-exploring the adaptation to unique chart characteristics, such as rich text elements. 
To fill the gap, we propose a visualization-referenced instruction tuning approach to guide the training dataset enhancement and model development. 
Specifically, we propose a novel data engine to effectively filter diverse and high-quality data from existing datasets and subsequently refine and augment the data using LLM-based generation techniques to better align with practical QA tasks and visual encodings. 
Then, to facilitate the adaptation to chart characteristics, we utilize the enriched data to train an MLLM by unfreezing the vision encoder and incorporating a mixture-of-resolution adaptation strategy for enhanced fine-grained recognition. 
Experimental results validate the effectiveness of our approach. 
Even with fewer training examples, our model consistently outperforms state-of-the-art CQA models on established benchmarks. 
We also contribute a dataset split as a benchmark for future research.
\revise{Source codes and datasets of this paper are available at \url{https://github.com/zengxingchen/ChartQA-MLLM}}.}

\teaser{
  \centering
  \includegraphics[width=\linewidth]{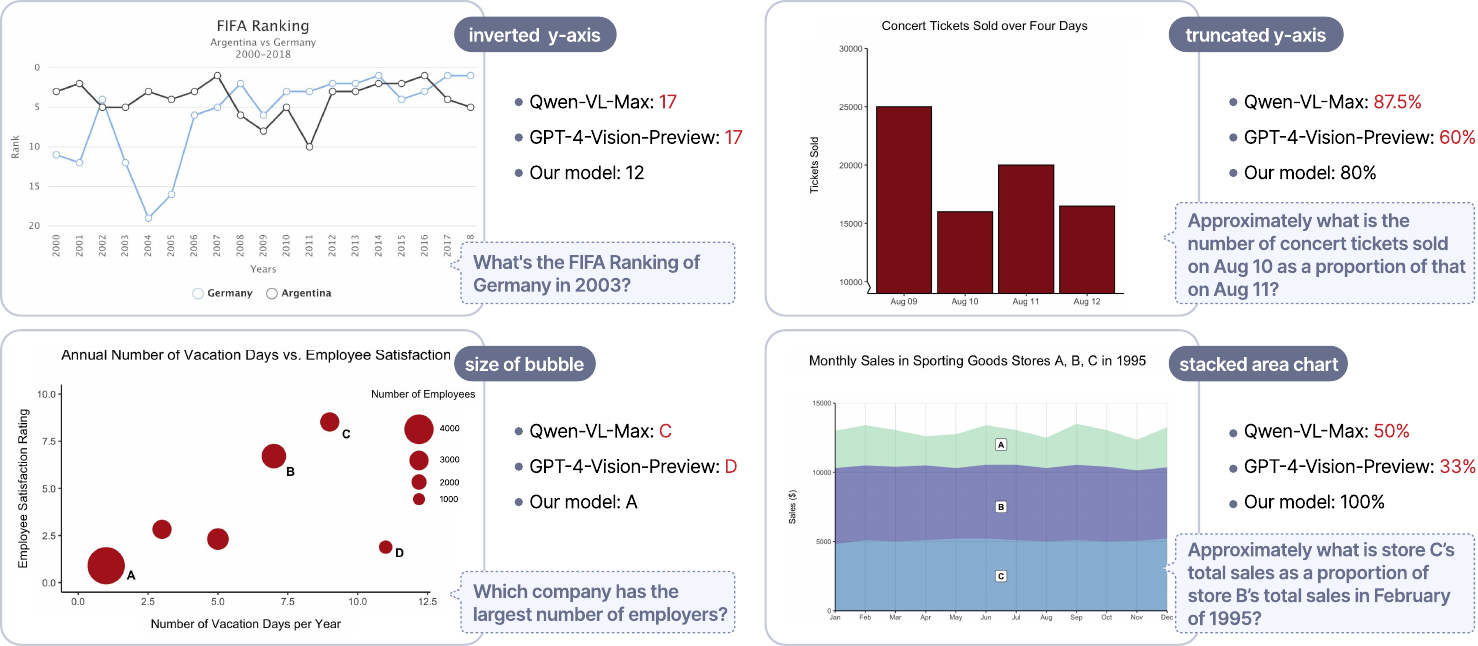}
  \caption{Comparison of our model with state-of-the-art MLLMs on chart question answering. Existing MLLMs often fail to understand visual mappings, such as inverted Y-axis, truncated axis, bubble sizing, and area stacking.
  In contrast, our model, trained with the visualization-referenced dataset we constructed, showcases a better understanding of visualization domain knowledge.}
  \label{fig:teaser}
}

\begin{document}



\maketitle
\section{Introduction}
Multimodal large language models (MLLMs), such as GPT4-Vision\cite{achiam2023gpt}, have made remarkable strides in understanding and interpreting natural images, enabling breakthroughs in various vision-language tasks (\eg, visual question answering\cite{antol2015vqa}). 
These models excel by aligning the image representation of pre-trained vision encoders with the powerful linguistic understanding of LLMs.
Thereby, MLLMs show great potential for visualization tasks that involve interpreting charts using natural language, such as chart question answering (CQA), chart summarization\cite{ye_2024_genai}, and chart reverse-engineering\cite{shi2024chartmimic}.
CQA poses intricate challenges, requiring both the comprehension of complex natural language, along with the recognition of information from charts and the reasoning ability to derive accurate answers\cite{hoque2022chart}.

Building MLLMs tailored for CQA necessitates high-quality training datasets and benchmarks. Recent research\cite{han2023chartllama, meng2024chartassisstant,liu2023mmc} in this field primarily focuses on scaling up training datasets that include charts, data tables, and QA pairs, employing manual labeling and data synthesis techniques.
These efforts have enhanced the performance of MLLMs in traditional CQA benchmarks\cite{masry2022chartqa,methani2020plotqa}.
However, bottlenecks have emerged, posing challenges for further improving performance and adapting to real-world scenarios.
Simply scaling up the training dataset without implementing quality control measures poses significant challenges in training efficiency and the feasibility of integrating these data into general MLLM training.

Recent research\cite{du2023makes} emphasizes the impact of different QA types on model performance, finding that reasoning-oriented\cite{du2023makes} and complexity-enhanced\cite{schwenk2022okvqa} instruction sets are particularly useful in improving the performance of MLLMs.
In the context of CQA, existing MLLMs for CQA encompass visual instructions in the format of {\tt <chart, question, answer>}.
The quality of the chart and question-answering (QA) pairs is pivotal for the effectiveness and generalizability of MLLMs.
However, the utilization of visual instruction data to enhance CQA remains largely under-explored, leaving unanswered questions about what makes good visual instructions and how to improve the dataset from the perspectives of visual instructions.

To address these inquiries, we conduct a comprehensive evaluation (Sect.~\ref{sec:empirical}) of MLLMs on CQA, aiming to pinpoint deficiencies and identify visual instructions that enhance MLLMs' performance.
The study utilizes the ChartQA dataset \cite{masry2022chartqa}, a widely adopted benchmark for CQA.
Through empirical analysis (Sect.~\ref{sssec:chartqa_bias}), we uncover notable distribution bias in both chart and QA pairs within the ChartQA dataset, as compared to practical datasets such as the Beagle image dataset\cite{battle2018beagle} and visual literacy assessment datasets\cite{lee2016vlat, ge2023calvi, pandey2023mini-vlat}.
Thorough experiments (Sect.~\ref{sssec:bias_impact}) uncover significant impacts of the distribution bias on MLLMs' performance in CQA, highlighting the necessity of incorporating more instructions for compositional and visual-compositional questions.
Ablation studies (Sect.~\ref{ssec:instruction_tuning_ablations}) further confirm that incorporating more reasoning-oriented QAs can significantly enhance model performance compared to including data retrieval QAs.

Drawing inspiration from the results, we introduce a novel data engine (Sect.~\ref{sec:data-engine}) to generate instruction-enhanced CQA datasets.
This engine comprises a data-filtering component (Sect.~\ref{ssec:data-filtering}), utilizing a classifier with fine-grained chart features to reveal distributions and filter existing chart datasets.
To mitigate the bias in the chart distributions and generate unavailable chart tasks, we further design a data generation component (Sect.~\ref{ssec:data-generation}) employing a chart space-guided data augmentation strategy to ensure the inclusivity of real-world possible charts.
We further enrich reasoning-oriented QAs for the generated charts, contributing to a new CQA dataset and benchmark (Sect.~\ref{ssec:benchmark}) that features a wider variety of chart types and more QAs with effective visual instructions.

Existing MLLMs, mostly relying on CLIP encoders trained on natural images, are not optimally suited for visualization charts due to inherent differences.
Recognizing the limitations, we develop a new MLLM (Sect.~\ref{sec:model}) that unfreezes the vision encoders in CLIP to better adapt to chart-specific features.
Our MLLM is trained using the newly curated CQA dataset with more effective visual instructions.
Additionally, we incorporate a mixture-of-resolution adaptation strategy\cite{luo2024feast} to enhance the fine-grained recognition capabilities of chart elements.
Quantitative experiments (Sect.~\ref{sec:eva}) demonstrate that even trained on a dataset with significantly less CQA data, our model consistently outperforms state-of-the-art CQA models on established benchmarks.

In summary, our contributions are three-fold:
\begin{itemize}
    \item An empirical study that identifies limitations of current MLLMs and ChartQA dataset and key factors (\ie, recognition and reasoning) that contribute to effective visual instructions for MLLMs' chart understanding.
    \vspace{-2mm}
    \item A novel data engine encompassing \textit{data filtering} and \textit{data generation}, producing a high-quality dataset and benchmark using visualization-referenced instruction tuning.
    \vspace{-2mm}
    \item An MLLM that outperforms existing open-source CQA models on existing CQA benchmarks and is comparable to the best commercial models on our proposed benchmark.
\end{itemize}

\section{Background of MLLMs}
Recently, LLMs\cite{touvron2023llama,brown2020language} have showcased powerful text generation and comprehension capabilities.
However, native LLMs live in the pure-text world and cannot process other common modalities such as images and videos, thereby limiting their application scope\cite{bai2023qwen}.
To break this limitation, a group of MLLMs (\eg, LLaVA\cite{liu2023llava}, Qwen-VL\cite{bai2023qwen}, and GPT4-Vision\cite{achiam2023gpt}) have emerged to endow LLMs with the ability to perceive and understand visual images.

\begin{figure}
    \centering
    \includegraphics[width=0.42\textwidth]{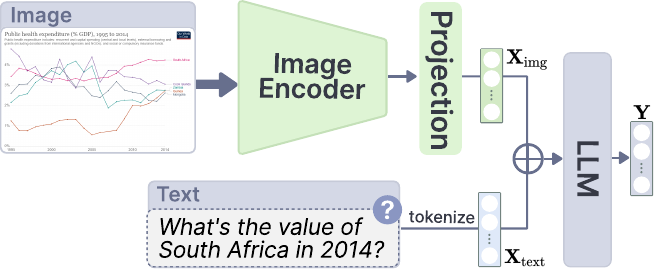}
    \caption{A typical architecture of MLLM, consisting of \textit{image encoder}, \textit{projector}, and \textit{LLM}.  \raisebox{-1.1pt}{\includegraphics[height=1.1em]{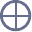}} represents the concatenation process of image  $\mathbf{X}_{\text {img}}$ and text tokens $\mathbf{X}_{\text {text}}$.}
    \label{fig:mllm-arch}
\end{figure}

Inspired by LLaVA\cite{liu2023llava}, current open-source MLLMs adopted a similar architecture to align the visual and textual features.
Figure~\ref{fig:mllm-arch} illustrates the typical MLLM architecture that comprises three modules: \textit{Vision Encoder}, \textit{Projection Layer}, and \textit{Large Language Model}.
Particularly, the \textit{Vision Encoder} (\eg, CLIP-Vit\cite{radford2021learning}) extracts a sequence of visual features from the input image. 
Then, the \textit{Projection Layer} (\eg, multiple linear layers\cite{liu2023llava} and querying transformer \cite{li2023blip}) transforms the visual features into the LLM word embedding space, resulting in compatible visual tokens $\mathbf{X}_{\text {img}}$ for the subsequent LLM (\eg, Vicuna-1.5\cite{zheng2024judging}).
Finally, the LLM processes the concatenated visual $\mathbf{X}_{\text {img}}$ and text tokens $\mathbf{X}_{\text {text}}$, \ie, $[\mathbf{X}_{\text {img}}, \mathbf{X}_{\text {text}}]$, and then autoregressively generates responses $\mathbf{Y}$.
Formally, the language model predicts the response $\mathbf{Y}=\left\{y_i\right\}_{i=1}^L$ conditioned on the multimodal input $\mathbf{X}_{\text {img}}, \mathbf{X}_{\text {text }}$, where $L$ means the number of tokens in the response. Therefore, the response is predicted by maximizing

\begin{equation}
p\left(\mathbf{Y} \mid \mathbf{X}_{\text {img}}, \mathbf{X}_{\text {text }}\right)=\prod_{i=1}^L p_\theta\left(y_i \mid \mathbf{X}_{\text {img}}, \mathbf{X}_{\text {text }}, y_{<i}\right),
\end{equation}
where $\theta$ is the trainable parameters. 

Despite the architectural harmonization, the biggest challenge in training generic MLLMs is collecting high-quality visual instruction data, \ie, $[\mathbf{X}_{\text {img}}, \mathbf{X}_{\text {text}}, \mathbf{Y}]$.
Visual instructions facilitate the alignment of the multimodal (\ie, language-image)  space, thus preserving and fusing the knowledge and abilities in the pre-trained vision encoder and LLM, empowering the MLLM with image-based conversation capabilities.
In a general form, visual instructions are composed of
{\tt <target image, text task description, text output>}, namely {\tt <chart, question, answer>} in CQA.

\section{Related Work}
\noindent \textbf{Vision-Language Models for Chart Understanding}.
Researchers have long been committed to developing vision-language (VL) models with strong capabilities in chart-related tasks (\eg, CQA and chart summarizing).
Previous works fall into two categories: 1) two-stage approaches that employ vision models to convert charts into data tables for subsequent processing with language models\cite{masry2022chartqa,liu2023deplot,do2023llms, cheng2023chartreader}; 
and 2) unified VL models that directly process and interpret the fused chart and text features in a single integrated phase\cite{masry2023unichart,liu2023matcha,meng2024chartassisstant}.

The two-stage pipeline struggles with preserving visual information (\eg, color and spatial location)\cite{liu2023deplot} when performing the chart-to-table transformation, which inherently limits their applicability to specific scenarios.
For unified models, Matcha\cite{liu2023matcha} integrates mathematical reasoning and chart data extraction tasks into a pre-trained generic VL model, Pix2Struct\cite{lee2023pix2struct}, thus excelling at CQA and chart summarizing. 
UniChart\cite{masry2023unichart} follows Matcha while collecting more data to undergo multitask instruction tuning for more chart-related tasks.
However, their limited language model performances pose challenges, especially in reasoning problems that necessitate numerical calculations\cite{masry2022chartqa}.

The advent of MLLMs has shifted the paradigm, achieving breakthroughs in visual question answering\cite{antol2015vqa}.
Notably, the open-source generic model Qwen-VL\cite{bai2023qwen} demonstrates superior performance over all specialized chart models in the ChartQA benchmark\cite{masry2022chartqa},
especially those posed by humans as opposed to machine-generated questions. 
Despite these advancements, our extensive empirical study has uncovered limitations in the current MLLMs' ability to handle real-world CQA tasks, especially those that fall outside of the training data distribution.
Rectifying these limitations necessitates the consideration of the visualization reference model\cite{card1999readings} when constructing training data, which elucidates the practical mapping process from raw data to final graphical representations. 
Accordingly, this study contributes to enhancing the performance of MLLMs in CQA by integrating knowledge from the visualization reference process into training data generation and augmentation.

\noindent \textbf{Enhancing Capabilities of MLLMs}.
The enhancement of MLLMs in specific scenarios, such as medicine images and text-dense images, can be categorized into two primary approaches: \textit{\textbf{model-centric}} works that aim to improve the performance and efficiency of vision encoders or projectors;
\textit{\textbf{data-centric}} works that try to improve the model performance by boosting the number and quality of training data.
In data-centric advancements, several studies employ powerful LLMs (\eg, GPT-4\cite{achiam2023gpt}) to generate various instruction-format VL tasks, like caption generation\cite{liu2023llava}.
Another line of studies has explored converting classical VL task datasets (\eg, COCO\cite{lin2014microsoft}) into an instruction-following format with pre-defined templates.
Within this context, to enhance chart comprehension, ChartLLaMA\cite{han2023chartllama} finetunes LLaVA with 160K instruction data generated by GPT-4.
Similarly, ChartAst\cite{meng2024chartassisstant} crawls a huge amount of tables from arXiv and then uses tables to generate charts for large-scale chart-to-table pre-training.
ChartAst also generates QA pairs based on the tables they collected.
Despite these efforts, the factors contributing to efficient instruction data for chart understanding are still unclear.

Our research seeks to investigate this gap with an empirical study that revisits the differences in improving model performance using different types of CQA task data.
The results underscore the significance of integrating complex chart reasoning questions, prompting us to develop a data engine enriched with real-world chart tasks.
Moreover, we have also made improvements to the model-centric side by tailoring the training methodology of base MLLMs, initially tailored for natural images, to suit visualization contexts.

\noindent \textbf{Visualization Datasets and Benchmarks}.
Datasets form the foundation of model training, and well-structured benchmarks help researchers evaluate and choose appropriate models for downstream tasks.
Specific to visualization scenarios, current benchmarks mainly focus on evaluating chart understanding performance via chart-to-table transformation\cite{masry2022chartqa, methani2020plotqa}, CQA tasks\cite{masry2022chartqa, methani2020plotqa}, and chart summarizing\cite{benny2023vistext, kantharaj2022chart2text, rahman2023chartsumm}.

ChartQA\cite{masry2022chartqa} and PlotQA\cite{methani2020plotqa} are representative of the QA datasets and benchmarks.
ChartQA features partially high-quality human-annotated QA pairs, while PlotQA offers a more voluminous collection of lower-quality items crafted using templates. 
Beyond QA tasks, VisText\cite{benny2023vistext} introduces a comprehensive benchmark, which incorporates multi-level and fine-grained chart labeling, covering aspects such as chart construction, summary statistics, relations, and complex trends. 
The primary strength of these datasets is their expansive size and the carefully crafted templates used for data generation. 
However, they have limitations, including a restricted range of chart types, the challenge of maintaining high-quality questions and answers, and a tendency to focus excessively on basic data retrieval from the charts.

In the visualization field, real-world image datasets like Beagle\cite{battle2018beagle}, VisImage\cite{deng2022visimages}, Vis30K\cite{chen2021vis30k}, multi-view~\cite{chen2021composition} and composite visualizations~\cite{deng_2023_composite, Giacomo_2024_comparative}, and dashboards~\cite{deng_2023_dashbot, survey_2024_dashboard}, together with practical QA benchmarks for visual literacy test\cite{lee2016vlat, pandey2023mini-vlat,ge2023calvi}, have been introduced. 
The challenge lies in converting them into high-quality instruction data due to sparse label annotations. 
Our research draws upon methodologies that utilize GPT to generate code-format charts and associated instruction data.
Specifically, we aim to guide the data generation process with the well-defined chart-task space\cite{lee2016vlat} to contribute a dataset encompassing the real-world spectrum of chart features and QA tasks, thereby improving current MLLMs' chart understanding capability.

\section{Empirical Study: Revisiting MLLMs for CQA}
\label{sec:empirical}

We conduct an empirical study to revisit the effectiveness of existing MLLMs for CQA, aiming to identify limitations and glean insights for further improvements.
The study is informed by the CQA leaderboard\footnote{\href{https://paperswithcode.com/task/chart-question-answering}{https://paperswithcode.com/task/chart-question-answering}} and recent research\cite{liu2024llavanext,bai2023qwen,achiam2023gpt}, where highlights that ChartQA\cite{masry2022chartqa} serves as the primary training and testing dataset for MLLM in chart understanding. 
ChartQA encompasses large-scale real-world charts sourced from online platforms, accompanied by data tables and both human-authored and machine-generated QA pairs.
Nevertheless, in-depth analyses are imperative to ensure that MLLMs exhibiting good benchmark performance on ChartQA can reliably transition to real-world scenarios.
In particular, this empirical study aims to address the following research questions:

\begin{itemize}
    \item \textit{RQ1: How can ChartQA be enhanced to reflect real-world scenarios better?}
    We aim to refine ChartQA to align more closely with real-world contexts.
    While the charts in ChartQA are sourced from online platforms, they do not encompass the entire spectrum of chart designs, as a recent study\cite{ye2022visatlas} identifies a biased distribution of online charts.
    Specifically, we will explore the diversity of chart design and QA pairs, both essential aspects for enhancing the effectiveness of CQA models.

    \item  \textit{RQ2: What makes effective visual instructions for CQA?}
    While QA pairs inherently serve as instruction data, they include various question types (\eg, \emph{data retrieval} and \emph{visual}).
    Exploring which specific QA features can better improve the effectiveness of visual instructions is notably under-explored. 
    Furthermore, previous studies\cite{liu2023deplot, masry2022chartqa} suggest that incorporating the chart-to-table translation task improves VL models' general chart understanding performance, while its effect within the context of MLLMs merits deeper investigation.
\end{itemize}

\subsection{Computational Analysis of ChartQA Dataset}
\label{ssec:chartqa-analsyis}
To address \emph{RQ1}, we conduct computational analyses of ChartQA's distribution in terms of chart and QA pairs.
We identify distribution bias by comparing them with practical charts and visual literacy data.
Subsequently, we assess the performance of various MLLMs on ChartQA and contrast these results with performances in real-world scenarios, emphasizing the impacts of distribution bias on model performance.

\subsubsection{Distribution Biases in Chart and QA Pairs}
\label{sssec:chartqa_bias}
\begin{figure}
    \centering
    \includegraphics[width=0.495\textwidth]{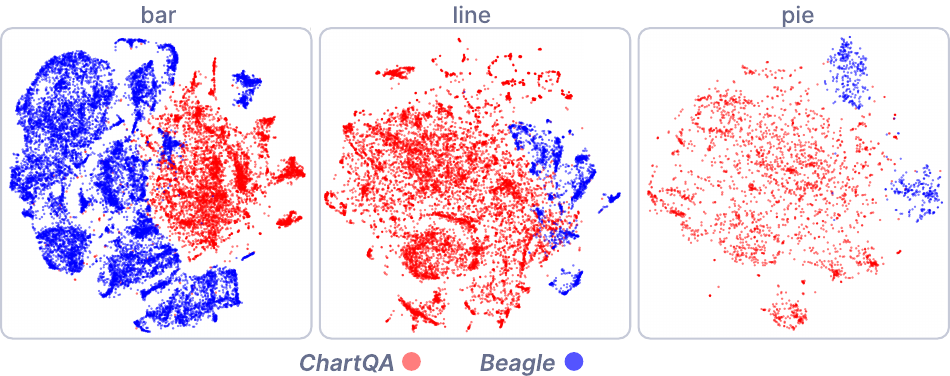}
    \caption{Comparison of chart distribution between ChartQA\cite{masry2022chartqa} and Beagle\cite{battle2018beagle}.}
    \vspace{-2mm}
    \label{fig:data-distribution}
\end{figure}

\begin{figure*}[thbp]
    \centering
    \includegraphics[width=0.995\textwidth]{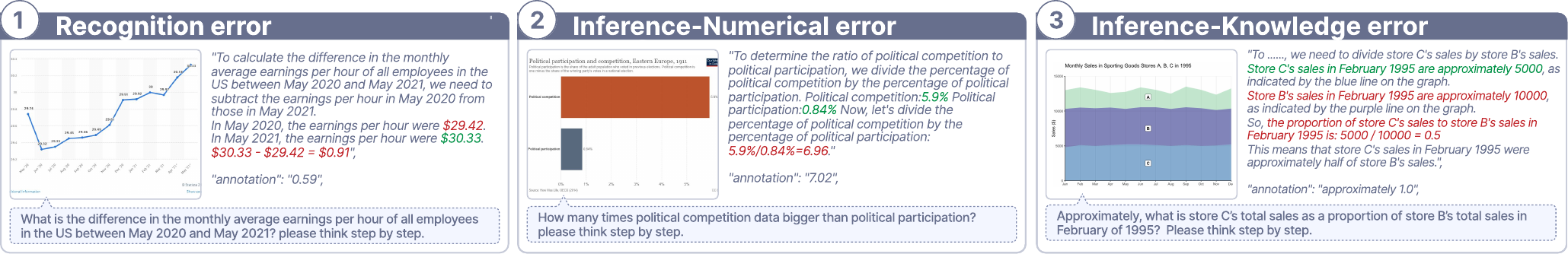}
    \vspace{-2mm}
    \caption{Three typical categories of failed reasons: \textit{recognition error}, \textit{numerical error during inference}, and \textit{knowledge error during inference}. False reasoning steps are colored in \redColor{red}, while correct steps are colored in \greenColor{green}. }
    \label{fig:failed-case}
\end{figure*}

\begin{table*}[thb]
    \centering
    \caption{Model performance on ChartQA and the collected visual literacy dataset. \emph{ChartQA-M} and \emph{ChartQA-H} refers to the machine-generated and human-annotated set, respectively. Particularly, \emph{ChartQA-H} is divided into four subcategories for fine-grained analysis.}
    \vspace{-2mm}
    \begin{tabular}{c|c|cccc|c}
        \hline
    \multirow{2}{*}{\textbf{Model}} & \multirow{2}{*}{\textbf{ChartQA-M}} & \multicolumn{4}{c|}{\textbf{ChartQA-H}}                                      & \multirow{2}{*}{\textbf{Literacy}} \\\cline{3-6}
                           &                            & \textbf{Data Retrieval} & \textbf{Compositional} & \textbf{Visual}     & \textbf{Visual-Compositional} &                           \\\hline
    LLaVA-1.6-13b          & $72.16 \%$                 & $70.28 \%$     & $30.24 \%$    & $66.10 \%$ & $10.81 \%$           & $22.13 \%$                \\
    LLaVA-1.6-34b          & $77.52 \%$                 & $71.49 \%$     & $44.35 \%$    & $71.19 \%$ & $35.14 \%$           & $35.11 \%$                \\
    Qwen-VL-Chat           & $85.36 \%$                 & $66.67 \%$     & $23.79 \%$    & $62.29 \%$ & $18.92 \%$           & $25.19 \%$                      \\
    Qwen-VL-Plus           & $70.32 \%$                 & $51.00 \%$     & $24.60 \%$    & $60.81 \%$ & $24.32 \%$           & $24.42 \%$                \\
    GPT-4-vision-preview           & $87.25\%$                  & $68.43\%$      & $25.96\%$     & $68.84\%$   & $20.85\%$           & $41.98 \%$               \\
    \hline
    \end{tabular}
    \label{tab:preliminary}
    \vspace{-3mm}
\end{table*}

\noindent{\textbf{Chart distribution.}} 
ChartQA primarily consists of bar, line, and pie charts sourced from online platforms.
These charts have similar visual styles (\eg, color themes) and lack the coverage of the diverse range of chart types such as area charts and scatterplots.
Moreover, even within their included chart types, there can be significant differences in fine-grained chart features compared to practical charts.
To investigate these differences, we utilize \textit{Beagle}\cite{battle2018beagle} as the control group to compare their distributions of chart features.
\textit{Beagle} crawls visualizations from the web using keyword searches and is considered to be comparatively diverse among available visualization datasets\cite{ye2022visatlas}, encompassing charts from various visualization tools and libraries (\eg, D3\cite{bostock2011d3} and Chartblocks\cite{chartblocks}).
Specifically, we use the pre-trained CLIP-Vit\cite{radford2021learning}, a commonly used vision encoder of MLLMs, to extract high-dimensional features from the images.
We then project the features into two dimensions with t-SNE\cite{van2008visualizing}.
Figure~\ref{fig:data-distribution} presents the projection results, which show the distribution bias within specific chart types.

Importantly, chart type serves as a broad categorization, limiting the utilization of fine-grained chart features in constructing visual instructions.
For instance, number annotations allow MLLMs to recognize and retrieve data directly.
In contrast, when numerical annotations are absent, MLLMs must approximate data values based on the axes and positions of visual elements, posing a significantly more complex task.
This highlights the need for considering fine-grained chart features when formulating CQA datasets, as elaborated in Sect.~\ref{ssec:data-filtering}.

\vspace{1mm}
\noindent{\textbf{QA pair distribution.}} 
ChartQA comprises two testing QA datasets: \emph{ChartQA-H} for human-authored QAs and \emph{ChartQA-M} for machine-generated QAs.
These QAs are categorized into \textit{data retrieval}, \textit{visual}, \textit{compositional}, and \textit{visual-and-compositional} types, as defined in \cite{kim2020answering}.
\begin{itemize}
    \item \emph{Data retrieval}: finding the value of the corresponding elements through the entity name in the chart.
    \vspace{-1mm}
    \item \emph{Visual}: leveraging visual channels, such as color identification, comparison between entities using visual attributes (\eg, which is rightmost, highest, or largest)
    \vspace{-1mm}
    \item \emph{Compositional}: requiring mathematical operations like sum, difference, and average.
    \vspace{-1mm}
    \item \emph{Visual-and-compositional}: blending of visual and compositional.
\end{itemize}

However, ChartQA does not annotate the question type for each QA pair, hindering the fine-grained accuracy analysis based on question types.
To address the issue, we manually labeled the questions in the \emph{ChartQA-H} and \emph{ChartQA-M} test sets, each containing 1250 QAs.
Statistics reveal that \textit{data retrieval} task (1035/1250) dominates the \emph{ChartQA-M} set.  
This distribution likely stems from the limited performance of the language model used for generation, which restricts \emph{ChartQA-M} to specific question templates.
In contrast, the human-authored \emph{ChartQA-H} set features a more diverse distribution, containing (251) \textit{data retrieval}, (476) \textit{visual}, (251) \textit{compositional}, and (272) \emph{visual-and-compositional} types.
The diversified distribution motivates us to conduct a more comprehensive evaluation of the model's chart understanding ability across different question types, as detailed in the subsequent section.

\subsubsection{Impacts of Distribution Bias}
\label{sssec:bias_impact}
We further study how the distribution bias identified in the above section affect the model performance.

\noindent \textbf{Models.}
Our selected  MLLMs include open-source models explicitly trained on \textit{ChartQA}: LLaVA-1.6-13b\cite{liu2024llavanext}, LLaVA-1.6-34b\cite{liu2024llavanext} and Qwen-VL-Chat\cite{bai2023qwen};
and mainstream commercial \revise{models}: Qwen-VL-Plus\cite{bai2023qwen}, and GPT-4-vision-preview\cite{gpt4-vision-preview}. 
The commercial \revise{models are} accessed through their official APIs.

\noindent \textbf{Evaluation metric.}
Following existing research\cite{masry2023unichart,masry2022chartqa,han2023chartllama}, we adopt the widely-used relaxed correctness~\cite{masry2022chartqa}, which requires exact matches for text responses but allows 5\% error for numerical responses.

\noindent \textbf{Prompt settings.}
The CQA evaluation requires the model to answer with a single word or short phrase. 
Following the LLaVA setup for short answers\cite{liu2023improvedllava}, we prompt the model with "\textit{Please answer with a single word or phrase}" for metric evaluation and "\textit{Please think step by step}" for zero-shot chain-of-thought (CoT)\cite{wei2022chain} to investigate the key and error steps in the model's reasoning process.

\noindent \textbf{Datasets.}
Besides \emph{ChartQA-H} and \emph{ChartQA-M} test sets, we have mixed QA pairs from studies on \textit{visual literacy}\cite{lee2016vlat, ge2023calvi, pandey2023mini-vlat}, resulting in the creation of a new dataset comprising 131 QA pairs.
Visual literacy QAs are designed to assess an individual's ability to read, comprehend, and interpret data visualizations. These are representative examples of real-world QAs covering most of the chart-task space\cite{lee2016vlat}.

\begin{table*}[thb] %
    \centering
    \caption{Results on \emph{ChartQA-H} test set with models trained on individual and different combinations of training datasets in ChartQA.}
    \vspace{-2mm}
    \begin{tabular}{lcccc}
    \hline 
    \textbf{Model} & \textbf{Data Retrieval} & \textbf{Compositional} &  \textbf{Visual} &  \textbf{Visual-Compositional} \\
    \hline 
    Baseline LLaVA-1.5 & 24.50\% & 9.27\% & 28.60\% & 13.51\%\\
    LLaVA-1.5 + \emph{ChartQA-H} & 32.93\% & 15.73\% & 47.25\% & 8.11\%\\
    LLaVA-1.5 + \emph{ChartQA-M} & 31.33\% & 10.08\% & 38.77\% & 8.11\%\\
    LLaVA-1.5 + \emph{Chart2Table} & 36.55\% & 9.68\% & 47.46\% & 13.51\%\\
    LLaVA-1.5 + \emph{ChartQA-H} \& \emph{ChartQA-M} & 43.37\% & 15.73\% & 51.91\% & 5.41\%\\
    LLaVA-1.5 + \emph{ChartQA-H} \& \emph{Chart2Table} & 42.17\% & 16.94\% & 51.91\% & 13.51\%\\
    LLaVA-1.5 + \emph{ChartQA-H} \& \emph{ChartQA-M} \& \emph{Chart2Table} & 48.59\% & 18.55\% & 54.66\% & 13.51\%\\
    \hline
    \end{tabular}
    \vspace{-1mm}
    \label{tab:ablation}
\end{table*}

\noindent \textbf{Result analysis.}
Table~\ref{tab:preliminary} presents the experimental results, showing that all MLLMs exhibit a performance disparity between \emph{ChartQA-M} and \emph{visual literacy}.
A plausible hypothesis is the uneven distribution of question types in \emph{ChartQA-M}.
To validate this hypothesis, we disaggregate the performances on different question types in \emph{ChartQA-H}.
The results unveil significant discrepancies among various question types.
Specifically, all models demonstrate high performances on \emph{data retrieval} and \emph{visual} questions, while their performances notably decline on \emph{compositional} and \emph{visual-compositional} questions.
Typically, \textit{data retrieval} and \textit{visual} questions mainly require the ability for chart recognition. 
In contrast, the \textit{compositional} questions need chart recognition followed by calculation and reasoning, heavily relying on MLLM's reasoning ability. 
This confirms the validity of the hypothesis.

To gain deeper insights into the underlying reasons for the issue, we examine the responses generated by MLLMs equipped with CoT.
Figure~\ref{fig:failed-case} illustrates three typical cases, highlighting deficiencies in three categories: recognition errors, inference errors for numerical calculations, and inference errors regarding chart knowledge.
Multiple factors contribute to these errors.
First, errors often occur for chart types common in \emph{visual literacy} but rare in ChartQA, such as stacked bar charts. 
Additionally, uncommon questions in ChartQA, such as accurately determining a range of data values, may lead MLLMs to struggle to identify the correct range.

\noindent \textbf{Summary.}
These insights highlight a crucial issue with ChartQA: while it includes a wide range of real-world images and QAs, biases in chart and QA distributions constrain its generalizability.
This emphasizes the need for a dataset incorporating a broader variety of chart types and QAs.
Such a dataset can potentially enhance MLLM's ability to tackle the complex challenges inherent in real-world scenarios.

\subsection{Instruction Tuning Ablations}
\label{ssec:instruction_tuning_ablations}
To address \emph{RQ2}, we design a series of ablation studies to examine the effect of different question types and chart-related tasks on CQA, aiming to identify effective visual instructions.

\subsubsection{Experiment settings}
\noindent \textbf{Backbone MLLM}: We select LLaVA-1.5\cite{liu2023improvedllava} as the baseline because its training data does not contain a specific chart dataset, making it easier to study the effect of different training data composition.
We follow the official fine-tuning settings of LLaVA-1.5, where we freeze the vision encoder and only update the parameters of the projector and the LLM. 
Specifically, we employ the Low-Rank Adaptation (LoRA)\cite{hu2022lora} strategy to train LLM to reduce the training workload.

\noindent \textbf{Dataset Control}:
Despite the biased chart distribution with ChartQA, we utilize it for instruction tuning ablation tests due to its suitability for examining how MLLMs learn from and react to specific data distributions.
In addition to \emph{ChartQA-H} and \emph{ChartQA-M}, each chart in ChartQA is associated with its data table, constituting a chart-to-table translation task, denoted as \emph{Chart2Table}.
\revise{Studies\cite{masry2022chartqa,liu2023deplot} reveal that \emph{Chart2Table} has the potential to enhance chart recognition capabilities, which justifies its inclusion in our ablation study.
Specifically, the instruction data for \emph{Chart2Table} are structured as 
{\tt <chart, "Please extract the underlying data table from the given chart", data table>}.
}

\noindent \textbf{Ablation Models}: We use the backbone MLLM without fine-tuning as the baseline.
Furthermore, we fine-tune the backbone model with individual and different combinations of \emph{ChartQA-H}, \emph{ChartQA-M}, and \emph{Chart2Table}, resulting in a total of six fine-tuned MLLMs.

\subsubsection{Results and Analysis}
Table~\ref{tab:ablation} shows the results of the ablation experiment of baseline and the fine-tuned MLLMs on different question types in \emph{ChartQA-H} test set.
Overall, models fine-tuned with more training data (individual \emph{vs.} combinations datasets) achieve higher accuracy.
Specifically, the inclusion of the human-generated \emph{ChartQA-H} dataset substantially enhances model performance across all question types.
In contrast, \emph{ChartQA-M} dataset is less effective and mainly improves data retrieval and visual questions.
This difference further underscores the limited impact of data retrieval questions for tackling CQA challenges and the critical role of diverse, reasoning-intensive questions over simple recognition questions.
\revise{Moreover, \emph{Chart2Table} serves as an accompanying effective instruction task if the data tables are available.}

In summary, enhancing MLLM's chart understanding necessitates focusing on diversity, especially in question types demanding reasoning, over expanding the volume of data retrieval-focused training examples.

\section{Data Engine}
\label{sec:data-engine}
\begin{figure*}[t]
    \centering
    \includegraphics[width=0.995\textwidth]{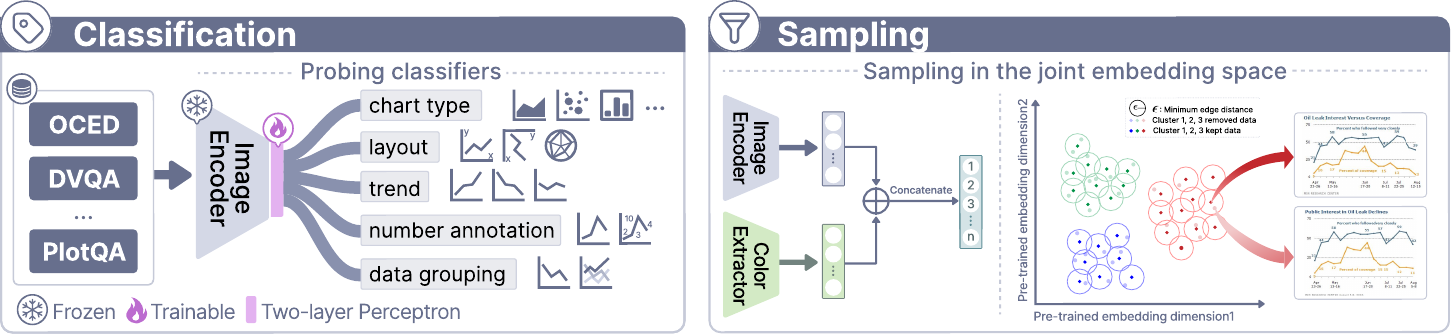}
    \vspace{-2mm}
    \caption{Illustration of the data filtering process, encompassing classification and sampling. Classification aims to investigate the distribution of existing datasets across key categorical attributes, including chart types, layout, trend, number annotations, and data grouping. Subsequently, we conduct sampling based on the fine category.}
    \label{fig:filtering}
    \vspace{-3mm}
\end{figure*}

\revise{Collecting all available CQA data for training an MLLM is inefficient and cannot address inherent distribution flaws.}
First, research has revealed the importance of data balance in training a generic MLLM\cite{cha2023honeybee}.
Without precise labeling, aggregating all data produces a massive dataset, causing learning inefficiency and training expensiveness of MLLMs\cite{he2024efficient}.
For instance, LLaVA\cite{liu2023llava} as a leading generic MLLM only requires 1223K
instruction data, whilst UniChart\cite{masry2023unichart} and ChartAssistant\cite{meng2024chartassisstant} use about 6900K and 39.4M chart-related instruction data. 
This disparity highlights the impracticality of incorporating all available chart data into generic MLLMs' training data.
\revise{Furthermore, our empirical study has demonstrated the distribution flaws in existing CQA data, underscoring the necessity of generating new data.}

To this end, we opt to design a data engine for a dataset of appropriate size while encompassing the real-world spectrum of chart features and QA tasks.
The data engine consists of two modules: \revise{\textit{\textbf{data filtering}} (Sect.~\ref{ssec:data-filtering}) for efficiently utilizing the existing data and also ensuring appropriate training cost;
and \textit{\textbf{data generation}} (Sect.~\ref{ssec:data-generation}) for optimizing the data distribution.}
Finally, we present the obtained \textit{\textbf{visualization-referenced dataset}} and \textit{\textbf{benchmark}} (Sect.~\ref{ssec:dataset-benchmark}).

\begin{table}[]
    \centering
    \caption{Statistics of existing datasets, only considering the training set if dataset splits (\ie, train-test) exist. 
    \revise{ Data counts consider the data tables and QA pairs associated with images.
    For example, a chart may be attached with its data table and two QA pairs, and then it is counted three times in total.
    }}
    \vspace{-2mm}
\begin{tabular}{lcc}
    \hline \textbf{Dataset} & \begin{tabular}{c} 
    \revise{\textit{\textbf{Chart tables}}}
    \end{tabular} & \begin{tabular}{c} 
    \revise{\textit{\textbf{Chart QA pairs}}}
    \end{tabular} \\
    \hline 
    Statista, OECD, OWID & 144,147 & 679,420 \\
    PlotQA & 155,082 & 2,414,359 \\
    Unichart & 189,792 & 2,218,468 \\
    Beagle & 3,972 & 51 \\
    ChartInfo & 1,796 & 21,949 \\
    VisText & 9,969 & 0 \\
    ExcelChart & 106,897 & 0 \\

    \hline Total existing & 611,655 & 5,334,247\\
    Filtered dataset & 69,418 & 68,223 \\
    \hline
    \label{tab:existing-datasets}
\end{tabular}
\vspace{-3mm}
\end{table}

\subsection{\revise{Data filtering}}
\label{ssec:data-filtering}
This module is designed to filter representative data from existing CQA datasets.
We first establish principles for what constitutes an appropriate chart distribution. 
Specifically, drawing on the taxonomy of the chart and corresponding task types outlined by~\cite{lee2016vlat} (see Table~\ref{tab:chart-task-space}), 
our methodology involves analyzing the distribution across the following:

\begin{itemize}
    \item chart types summarized in visualization literacy papers\cite{lee2016vlat,ge2023calvi} (see Table~\ref{tab:chart-task-space} for details);
    \vspace{-1mm}
    \item fine-grained chart attributes identified in visualization retrieval tasks\cite{xiao2023wytiwyr}, \eg, color, trends and layouts; and
    \vspace{-1mm}
    \item chart attributes that significantly affect MLLMs' chart understanding, \ie, number annotations (existent or absent) and data grouping (single or multiple). 
\end{itemize}

\noindent \revise{Given that studies have shown common pre-trained visual feature extractors (\eg, CLIP-Vit\cite{radford2021learning}) are not sensitive to fine-grained chart attributes\cite{xiao2023wytiwyr}, 
conventional filtering approaches that sample data in the pre-trained feature space lead to inhomogeneity in these attributes.
Additionally, most existing datasets lack detailed annotations beyond coarse-grained chart types (\eg, bar, line, and pie), posing challenges for stratified sampling.
To mitigate this issue, we construct classifiers to learn those attributes in a supervised manner and then perform stratified sampling based on the labels predicted by the classifiers.
}

\subsubsection{Image Classifier}
As shown in Figure~\ref{fig:filtering} \raisebox{-2pt}{\includegraphics[height=1em]{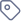}}, we build probing classifiers (\ie, two-layer perceptron) based on the frozen ConvNeXt\cite{woo2023convnext} backbone to accurately assess the distribution of these fine-grained chart types and attributes.
We collect training data sourced from \cite{xiao2023wytiwyr} alongside a manually collected subset.
To mitigate the issue of lacking some attribute annotations, we manually labeled each image for missing attributes.
Due to the unbalanced nature of chart type and attribute distribution (\eg, number annotations), we choose to use \emph{focal loss}\cite{lin2017focal} as the loss function, designed to focus on unbalanced image type.
Focal loss is defined as:
$$
\mathcal{L}_{\mathrm{FL}}\left(p_t\right)=-\alpha_t\left(1-p_t\right)^\gamma \log \left(p_t\right),
$$
where $p_t \in[0,1]$ represents the estimated probability of class $t, \alpha_t$ represents the scaling factor, and $\gamma$ represents the modulating factor. Among them, $\alpha_t$ is set by inverse class frequency. Thus, learning parameters tend to contribute to classes with fewer samples, and $\gamma$ assists in up-weighting the loss assigned to poorly-classified examples, avoiding the possibility that the amount of well-classified samples dominates the training process.
We empirically compare several design alternatives of backbone models (\eg, CLIP-Vit\cite{radford2021learning} and ResNet50\cite{he2016deep}) and trainable modules (\eg, linear probe\cite{radford2021learning}) in Table~\ref{tab:classifier-acc}.

Note that not all chart types possess the same set of fine-grained attributes. 
For instance, pie charts do not exhibit a trend attribute, so the trend classifier training will not consider pie charts.
\revise{We leverage the trained classifiers to label our collected existing data, providing clear inspections of the chart attributes and laying the foundation for data balancing in the subsequent sampling.}

\subsubsection{Image Sampling and Instruction Data Sampling}
\revise{
Figure~\ref{fig:filtering}  \raisebox{-2pt}{\includegraphics[height=1em]{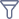}} illustrates the image sampling process.
We employ the CLIP-Vit\cite{radford2021learning} and a color extractor \cite{color-extraction} to extract the overall feature and color feature of each image and then concatenate the two feature vectors to formulate a joint embedding space.
} Inspired by Bunny\cite{he2024efficient} and SemDeDup\cite{abbas2023semdedup}, 
we cluster images into $k$ clusters via $k$-means within the joint embedding space, aiming to group charts with similar features.
\revise{To ensure chart attribute balancing, we incorporate stratified sampling within each cluster. 
Specifically, we create strata within each cluster according to predicted chart attributes and further perform sampling in each stratum.
We identify duplicates by constructing an undirected graph, where edges connect image pairs with cosine similarity above a specified threshold $\mathcal{\varepsilon}$, indicating high feature similarity.
We streamline the process by directly retaining only the image with the lowest cosine similarity to the stratum's centroid from each set of semantic duplicates,}
thereby effectively reducing dataset size while preserving diversity.
Finally, we manually adjust $\mathcal{\varepsilon}$ to obtain approximately 69K charts, ensuring an appropriate training cost.

For instruction data, Sect.~\ref{ssec:instruction_tuning_ablations} summarizes the effects of different components of existing datasets.
Drawing on the insights from the empirical study, we keep the table data of all sampled images for \emph{Chart2Table} task and further sample numerical and visual reasoning questions in their attached QA pairs.

\subsection{Data Generation}
\label{ssec:data-generation}

\begin{figure*}[htbp]
    \centering
    \includegraphics[width=0.995\textwidth]{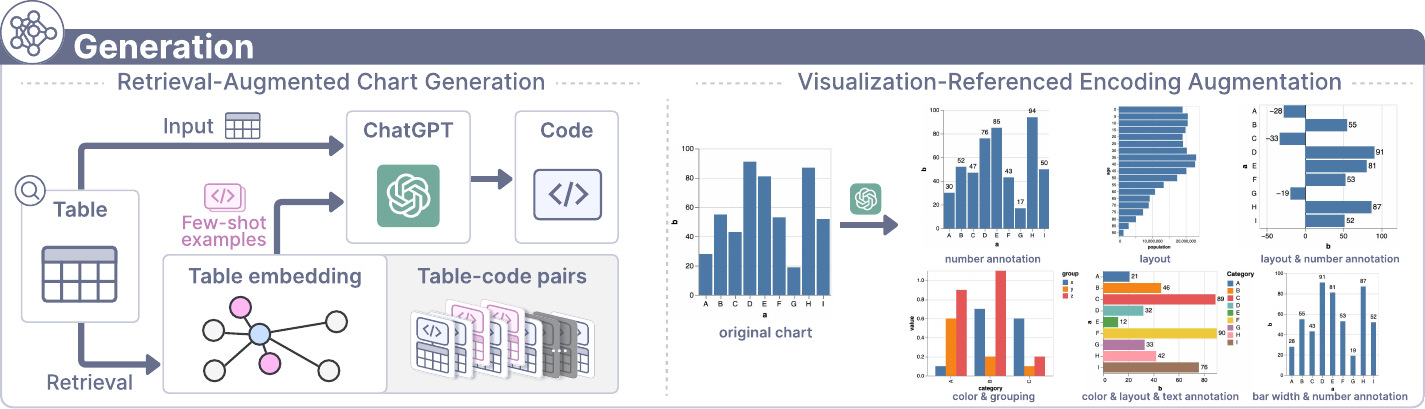}
    \vspace{-2mm}
    \caption{Data generation pipeline. First, we conduct retrieval-augmented chart generation with a set of table-code pairs we collected. This results in a collection of images distributed evenly in the real-world chart space. Then, we conduct visualization-referenced encoding augmentation for each seed chart to further enrich the dataset's size and diversity.}
    \vspace{-4mm}
    \label{fig:generation-pipeline}
\end{figure*}

This module is designed to generate a dataset encompassing real-world chart types and QA tasks, \revise{thus alleviating the distribution bias issue of existing datasets.} 
Specifically, we refer to the chart-task space types summarized in visualization literacy research\cite{lee2016vlat} (see Table~\ref{tab:chart-task-space}).

With collected tables, former works have explored generating charts and QA pairs using LLMs \cite{han2023chartllama, meng2024chartassisstant}.
However, they overlook potential quality distortions arising from the instability inherent in language model outputs,
nor has it considered guiding the generative process through an informed understanding of the chart space.
We harness LLMs' in-context learning ability to follow the visualization reference process\cite{card1999readings}, ensuring the variety of the resulting charts and thus covering the chart space. 
Figure~\ref{fig:generation-pipeline} outlines our chart generation pipeline, which encompasses two phases: \textit{Retrieval-Augmented Chart Generation} and \textit{Visualization-Referenced Encoding Augmentation}.
\revise{Templates of prompts constructed for LLM input in this section can be found in Supplementary Section S1.}

\subsubsection{Collection and Expansion of Seed Charts}
Generating charts begins with aggregating a diverse and high-quality set of seed examples that cover a wide representation of styles and chart types.
These examples are table-code pairs collected from a variety of authoritative chart libraries, such as Vega-Lite\footnote{\href{https://vega.github.io/vega-lite/examples}{https://vega.github.io/vega-lite/examples}}, 
Matplotlib\footnote{\href{https://matplotlib.org/stable/gallery}{https://matplotlib.org/stable/gallery}}, Seaborn\footnote{\href{https://seaborn.pydata.org/examples}{https://seaborn.pydata.org/examples}}, and ECharts\footnote{\href{https://echarts.apache.org/examples}{https://echarts.apache.org/examples}}. 
Moreover, we collect high-quality table-code pairs from previous studies\cite{ko2023natural} and handpick select examples from the web.
To further expand our seed examples, we also gather high-quality table data from various sources\cite{luo2021synthesizing, ye2022visatlas}.
\revise{Notably, charts filtered from existing datasets are not used here, as most of them are not in code format.}

As depicted in Figure~\ref{fig:generation-pipeline} (left), the expansion process employs the retrieval-augmented generation (RAG) method\cite{lewis2020retrieval},
which enhances the accuracy and quality of the generated charts by providing LLMs with contextually relevant examples during the generation process.
To implement this, we first extract table features to identify and match each collected table with the most similar tables among existing table-code pairs. 
Specifically, following visualization recommendation research\cite{hu2019vizml, li2021kg4vis, wang2023llm4vis}, we extract 30 cross-column data features that capture the relationships between columns and 81 single-column data features that quantify the properties of each column. 
These features allow us to represent the table features of the seed examples in a vector space, enabling the retrieval of nearest neighbors based on cosine similarity.
When constructing prompts for seed chart expansion, the corresponding codes of these matched seed examples serve as few-shot examples alongside the new tables. 

\subsubsection{Enhancement Through Visual Mapping Variations}
To broaden the collection of seed examples, we leverage the LLM to introduce variations in the visual mappings or encodings of charts.
This phase follows the visualization reference process\cite{card1999readings} and is crucial to encompass a broader array of possible chart presentations and to align with the diverse distributions of real-world data. 
We guide the LLM by specifying which visual mappings each chart type can adopt and incorporating instances featuring diverse visual encodings in the input context for reference. 
For instance, as illustrated in Figure~\ref{fig:generation-pipeline} (right), the LLM is instructed to modify chart elements like number annotations, groupings, and bar widths and to truncate or invert axes in bar charts. 
This approach facilitates generating a richer collection of table-code pairs by varying data-related encodings, such as the height of bars. 
The modified data tables are correspondingly recorded.

\subsubsection{Generation of Question and Answer Pairs}
\revise{We further generate QA pairs based on the enriched set of table-code pairs, which are expected to be accurate and cover the chart-task space.
Specifically, for each type of chart, we employ the LLM to generate Q\&A pairs by prompting it with tables for numerical information, code for encoded visual information, and the corresponding chart-task space as context. 
We also require the LLM to classify generated Q\&A pairs with category labels following the chart-task taxonomy (\eg, \emph{data retrieval} and \emph{find extremum}), balancing the distribution of different tasks. 
During the generation process, we randomly select some QA pairs for manual checking and ensure they are as accurate as expected.
}

\noindent \textbf{Reasoning process.} 
Recent studies show that unnecessary step-by-step training annotation leads to downgraded generalizability and instruction-following 
ability.
For simple questions (\eg, data retrieval of bar charts), the reasoning process does not provide useful information compared to single-word answers.
We only attach reasoning processes to questions that need numerical calculations and visual references. 
For visual references, we mainly consider visual channels that are less used in former work, such as the point area of the bubble chart and the truncated or inverted axis of line charts.

\subsection{Visualization-referenced Dataset and Benchmark}
\label{ssec:benchmark}

\setlength{\tabcolsep}{1.5pt}
\begin{table*}[htbp]
    \centering
    \caption{The chart-task space of our dataset, which is summarized by the visual literacy research \emph{VLAT}\cite{lee2016vlat}.}
    \vspace{-2mm}
    \resizebox{\textwidth}{!}{
    \begin{tabular}{c|ccccccccc|c}
    \hline
    Visualization           &                                                          &                                                           &                                                             &                                                                       & Visualization Task                                         &                                                           &                                                                           &                                                            &                                                                                            & Note of $\mathrm{X}^{\dagger}$                                                            \\ \hline
                            & \begin{tabular}[c]{@{}c@{}}Data\\ Retrieval\end{tabular} & \begin{tabular}[c]{@{}c@{}}Find \\  Extremum\end{tabular} & \begin{tabular}[c]{@{}c@{}}Determine \\  Range\end{tabular} & \begin{tabular}[c]{@{}c@{}}Characterize \\  Distribution\end{tabular} & \begin{tabular}[c]{@{}c@{}}Find \\  Anomalies\end{tabular} & \begin{tabular}[c]{@{}c@{}}Find \\  Clusters\end{tabular} & \begin{tabular}[c]{@{}c@{}}Find \\  Correlations/ \\  Trends\end{tabular} & \begin{tabular}[c]{@{}c@{}}Make\\ Comparisons\end{tabular} & ETC                                                                                        &                                                                                      \\ \hline
    Line Chart              & \multicolumn{1}{c|}{X}                                 & \multicolumn{1}{c|}{X}                                  & \multicolumn{1}{c|}{X}                                    & \multicolumn{1}{c|}{}                                                 & \multicolumn{1}{c|}{}                                      & \multicolumn{1}{c|}{}                                     & \multicolumn{1}{c|}{X}                                                  & \multicolumn{1}{c|}{X}                                   &                                                                                            &                                                                                      \\ \hline
    Bar Chart               & \multicolumn{1}{c|}{$\mathrm{X}^{\dagger}$}                             & \multicolumn{1}{c|}{X}                                  & \multicolumn{1}{c|}{X}                                    & \multicolumn{1}{c|}{}                                                 & \multicolumn{1}{c|}{}                                      & \multicolumn{1}{c|}{}                                     & \multicolumn{1}{c|}{}                                                     & \multicolumn{1}{c|}{$\mathrm{X}$}                               &                                                                                            &  \\ \hline
    Stacked Bar Chart       & \multicolumn{1}{c|}{$\mathrm{X}^{\dagger}$}                             & \multicolumn{1}{c|}{X}                                  & \multicolumn{1}{c|}{X}                                    & \multicolumn{1}{c|}{}                                                 & \multicolumn{1}{c|}{}                                      & \multicolumn{1}{c|}{}                                     & \multicolumn{1}{c|}{}                                                     & \multicolumn{1}{c|}{$\mathrm{X}^{\dagger}$}                               &                                                                                            & \begin{tabular}[c]{@{}c@{}}${\dagger}$  Both Absolute Value \\  and Relative Value\end{tabular} \\ \hline
    100\% Stacked Bar Chart & \multicolumn{1}{c|}{$\mathrm{X}^{\dagger}$}                             & \multicolumn{1}{c|}{$\mathrm{X}^{\dagger}$}                              & \multicolumn{1}{c|}{}                                       & \multicolumn{1}{c|}{}                                                 & \multicolumn{1}{c|}{}                                      & \multicolumn{1}{c|}{}                                     & \multicolumn{1}{c|}{}                                                     & \multicolumn{1}{c|}{$\mathrm{X}^{\dagger}$}                               &                                                                                            & ${\dagger}$   Only Relative Value                                                         \\ \hline
    Pie Chart               & \multicolumn{1}{c|}{$\mathrm{X}^{\dagger}$}                             & \multicolumn{1}{c|}{$\mathrm{X}^{\dagger}$}                              & \multicolumn{1}{c|}{}                                       & \multicolumn{1}{c|}{}                                                 & \multicolumn{1}{c|}{}                                      & \multicolumn{1}{c|}{}                                     & \multicolumn{1}{c|}{}                                                     & \multicolumn{1}{c|}{$\mathrm{X}^{\dagger}$}                                 &                                                                                            & ${\dagger}$   Only Relative Value                                                         \\ \hline
    Histogram               & \multicolumn{1}{c|}{$\mathrm{X}^{\dagger}$}                             & \multicolumn{1}{c|}{$\mathrm{X}^{\dagger}$}                              & \multicolumn{1}{c|}{}                                       & \multicolumn{1}{c|}{X}                                              & \multicolumn{1}{c|}{}                                      & \multicolumn{1}{c|}{}                                     & \multicolumn{1}{c|}{}                                                     & \multicolumn{1}{c|}{$\mathrm{X}^{\dagger}$}                               & \begin{tabular}[c]{@{}c@{}}Identify the \\  Characteristic of Bins\end{tabular}            & ${\dagger}$   Only Derived Value                                                          \\ \hline
    Scatterplot             & \multicolumn{1}{c|}{X}                                 & \multicolumn{1}{c|}{X}                                  & \multicolumn{1}{c|}{X}                                    & \multicolumn{1}{c|}{X}                                              & \multicolumn{1}{c|}{X}                                   & \multicolumn{1}{c|}{X}                                  & \multicolumn{1}{c|}{X}                                                  & \multicolumn{1}{c|}{X}                                   &                                                                                            &                                                                                      \\ \hline
    Area Chart              & \multicolumn{1}{c|}{X}                                 & \multicolumn{1}{c|}{X}                                  & \multicolumn{1}{c|}{X}                                    & \multicolumn{1}{c|}{}                                                 & \multicolumn{1}{c|}{}                                      & \multicolumn{1}{c|}{}                                     & \multicolumn{1}{c|}{X}                                                  & \multicolumn{1}{c|}{X}                                   &                                                                                            &                                                                                      \\ \hline
    Stacked Area Chart      & \multicolumn{1}{c|}{$\mathrm{X}^{\dagger}$}                             & \multicolumn{1}{c|}{X}                                  & \multicolumn{1}{c|}{X}                                    & \multicolumn{1}{c|}{}                                                 & \multicolumn{1}{c|}{}                                      & \multicolumn{1}{c|}{}                                     & \multicolumn{1}{c|}{X}                                                  & \multicolumn{1}{c|}{$\mathrm{X}^{\dagger}$}                               &                                                                                            & \begin{tabular}[c]{@{}c@{}}${\dagger}$  Both Absolute Value \\  and Relative Value\end{tabular} \\ \hline
    Bubble Chart            & \multicolumn{1}{c|}{X}                                 & \multicolumn{1}{c|}{X}                                  & \multicolumn{1}{c|}{X}                                    & \multicolumn{1}{c|}{X}                                              & \multicolumn{1}{c|}{X}                                   & \multicolumn{1}{c|}{X}                                  & \multicolumn{1}{c|}{X}                                                  & \multicolumn{1}{c|}{X}                                   &                                                                                            &                                                                                      \\ \hline                       
    Treemap                 & \multicolumn{1}{c|}{$\mathrm{X}^{\dagger}$}                             & \multicolumn{1}{c|}{$\mathrm{X}^{\dagger}$}                              & \multicolumn{1}{c|}{}                                       & \multicolumn{1}{c|}{}                                                 & \multicolumn{1}{c|}{}                                      & \multicolumn{1}{c|}{}                                     & \multicolumn{1}{c|}{}                                                     & \multicolumn{1}{c|}{$\mathrm{X}^{\dagger}$}                               & \begin{tabular}[c]{@{}c@{}}Identify the Hierarchical \\  Structure of Dataset\end{tabular} & ${\dagger}$   Only Relative Value                                                         \\ \hline
    \end{tabular}
    }
    \vspace{-3mm}
    \label{tab:chart-task-space}
\end{table*}

\label{ssec:dataset-benchmark}
\noindent \textbf{Overview of Dataset}.
Our generated dataset comprises 11 chart types and 8 task categories, as outlined in visual literacy research \cite{lee2016vlat, ge2023calvi, pandey2023mini-vlat}. 
Table~\ref{tab:chart-task-space} illustrates the chart-task space. 
The generated dataset includes 10,385 table-chart pairs and 51,245 chart-QA pairs.
\revise{By integrating the generated dataset with the filtered dataset (shown in Table~\ref{tab:existing-datasets}), we ultimately produce a dataset of 199K, which includes 80K table-chart pairs and 119K chart-QA pairs.}
In contrast to earlier datasets that relied on single sources or template-driven designs, our approach effectively combines our generated dataset with existing datasets derived from a wide range of real-world and synthetic sources. 
This integration features a diverse array of visual encodings, referencing real-world charts from multiple sources.
We have made deliberate efforts to ensure an equitable distribution of chart types and QA pairs, with a particular emphasis on underrepresented chart types and question types in current datasets, such as range determination and distribution characterization, to provide a more comprehensive and balanced resource.
\revise{Detailed chart and QA examples can be found in Supplementary Section S2.}

\noindent \textbf{Benchmark.}
To establish a benchmark covering the chart-task space, we meticulously curated an additional 368 table-chart pairs and 736 chart-QA pairs highly representative of our dataset. 
We focus on achieving diversity within each chart category, selecting charts with various visual encodings, and maintaining coverage into chart sub-types. 
Additionally, we consider the complexity of the question, aiming for a wide representation in both the number of entities and the range of quantities presented.

\noindent \textbf{Metrics.}
Our dataset is designed to reflect authentic scenarios encountered in chart-based question answering, and for that purpose, we have chosen to utilize the GPT for evaluation\cite{liu2023gpteval}. 
This manner is suitable for our benchmark as it can accommodate a wide range of answer formats, including ambiguous or long texts. 
The GPT accuracy metric compares textual responses to a standard expected answer, ensuring a match based on semantic equivalence.
For numerical responses, we allow a tolerance level of 5\%, which is consistent with former works\cite{masry2022chartqa}. 
However, this error margin is subject to adjustment in specific scenarios where it is inappropriate. For instance, in cases involving years or countable quantities, precision is crucial, and as such, absolute accuracy is demanded, with no error margin permitted.
\section{Model}
\label{sec:model}
To enhance MLLMs' chart comprehension in real-world contexts, we consider two design improvements in both the model architecture and its training.
Particularly, we adopt a mixture-of-resolution adaptation strategy\cite{luo2024feast} for enhanced fine-grained recognition (Sect.~\ref{ssec:model-arch}).
Moreover, for a better representation of the chart's visual feature,  we unfreeze the vision encoder during training and utilize the visualization-referenced dataset described in Sect.~\ref{sec:data-engine} for training (Sect.~\ref{ssec:model-training}).

\subsection{Model Architecture}
\label{ssec:model-arch}
\begin{figure}[htbp]
    \centering
    \includegraphics[width=0.49\textwidth]{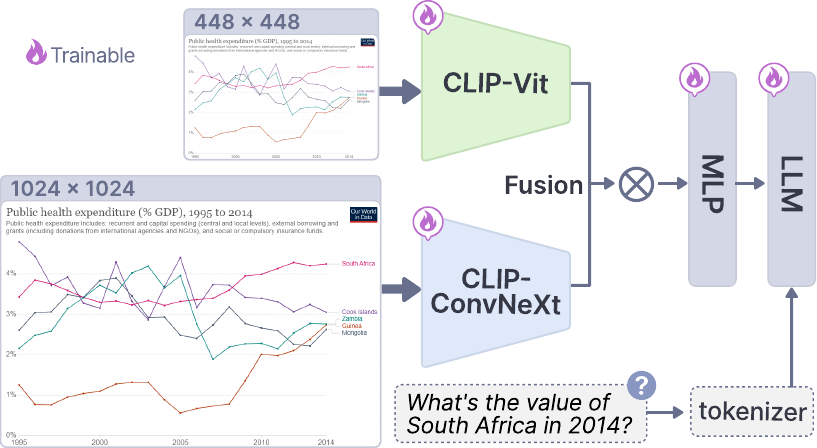}
    \caption{Architecture of the MLLM adopted in our work. High-resolution and normal-resolution features of the input image are fused to facilitate the efficient recognition of fine-grained features.
    During the training phase, vision encoders are unfreezed to enable the adaptation to chart characteristics.}
    \label{fig:our-arch}
    \vspace{-2mm}
\end{figure}

\noindent 
\textbf{Base model}. We use LLaVA-1.5\cite{liu2023llava} as the base model architecture, which employs CLIP-Vit-334px as the vision encoder, two-layer MLP as the projector, and the Vicuna-13B\cite{zheng2024judging} as the LLM.

\noindent \textbf{High-resolution input by mixing vision encoders with tokens compression.} 
The employed two-layer MLP effectively connects the feature space of vision encoders and LLMs but results in visual tokens that positively correlate with the image resolution.
For instance, CLIP-Vit-L-14 results in 5,329 tokens for a 1,022×1,022 resolution image, as each token corresponds to a 14×14 image patch\cite{luo2024feast}, which is computationally expensive for MLLMs.
The popular query-based strategy for implementing high-resolution input, QFormer\cite{li2023blip}, requires large-scale pre-training to achieve vision-language alignments, which is impractical for visualization scenarios as high-quality data is scarce.
Therefore, we adopt a resolution-adaptation strategy\cite{luo2024feast} to improve the resolution while supporting training on normal scale data.
The strategy embeds high-resolution features into the low-resolution features via adapters, thus reducing visual feature tokens.
Specifically, following the settings of LLaVA-HR\cite{luo2024feast}, we integrate CLIP-ViT-L\cite{radford2021learning} and CLIP-ConvNeXt\cite{liu2022convnet} as a mixture of vision encoders and then mix their features with an adaptation strategy, thus maintaining control over the length of the visual token sequence. 
The resolutions of ViT and CNN are set to 448×448 and 1,024×1,024, respectively.

\subsection{Training Settings}
\label{ssec:model-training}
\noindent \textbf{Unfreezing the vision encoders.}
Former works\cite{liu2023llava,han2023chartllama,liu2023mmc} choose to freeze the vision encoder in the whole training process as the pre-trained CLIP is already good at capturing features of natural images.
Their MLLM training target is aligning the features extracted by CLIP to the LLM embedding space by tuning the projector and LLM.
\revise{However, former research\cite{xiao2023wytiwyr} has found that CLIP performs much worse in visualization images, as its pre-training corpus has a relatively small amount of charts with coarse annotation, leading to limited chart recognition ability without further tuning.}
Unfreezing CLIP's parameters enables better adaptation to chart features and improves the overall performance of MLLM's chart understanding for improved chart recognition ability.

\noindent \textbf{Training data.}
We skip the pre-training process and directly leverage the initial projector weight of LLaVA-HR\cite{luo2024feast} to conduct instruction tuning.
Our study aims to improve general MLLM's chart comprehension ability.
Therefore, our training data consists of two parts: the 665K original instruction tuning data of LLaVA-1.5 and the \revise{199K} chart-related data described in Sect.~\ref{ssec:dataset-benchmark}.

\noindent \textbf{Hyperparameters settings.}
AdamW\cite{kingma2014adam} is used as the optimizer, and the learning rate (LR) and the global batch size are set to 2e-5 and 128, respectively.
The training epoch is set to 1.
The LR scheduler is cosine decay, with a warmup ratio of 0.03.
The training is running on 16×NVIDIA A800 for approximately 19 hours.

\section{Evaluation}
\label{sec:eva}
We first present the accuracy of our classifier to show its effectiveness in measuring the chart distribution.
Then, we compare our model with previous works in traditional benchmarks and our benchmark.
Also, we provide a group of ablation studies to show the effectiveness of our data engine.

\subsection{Chart Attributes Classification}
We compare the performance of classifiers with different backbones (\ie, ResNet50\cite{he2016deep}, ConvNeXt\cite{liu2022convnet}, and CLIP-Vit\cite{radford2021learning}) and trainable modules (linear probe\cite{radford2021learning} and two-layer MLP with focal loss\cite{lin2017focal}). 
Table~\ref{tab:classifier-acc} lists each attribute's macro \emph{F1} score, showing that ConvNeXt consistently outperforms other backbones.
Therefore, we adopt the ConvNeXt designated focal loss, which consistently performs well.

\begin{table}[t]
    \centering
    \caption{Test classification performance of our proposed model on the dataset, comparing design choices about backbone models and loss functions. 
    \textbf{Bold} numbers indicate the highest metric value.}
    \vspace{-2mm}
\begin{tabular}{lccccc}
    \hline Models & \begin{tabular}{c} 
    \textit{Chart}
    \\ 
    \textit{
    Type}
    \end{tabular}&  \begin{tabular}{c} 
    \textit{Number}
    \\ 
    \textit{Annotation}
    \end{tabular} & \begin{tabular}{c} 
    \textit{Data}
    \\ 
    \textit{Grouping}
    \end{tabular} & \begin{tabular}{c} 
    \textit{Trend}
    \end{tabular} & \begin{tabular}{c} 
    \textit{Layout}
    \end{tabular} 
    \\
    \hline 
    ResNet50+Linear Probe & 90.4 & 86.8  & 85.8 &  75.2 & 95.1\\
    ResNet50+Focal Loss & 90.6 & 84.3 & 87.0 & 72.9 & 94.1 \\
    CLIP-Vit+Linear Probe & 93.0 & 87.6  & 92.7 &  70.8 & 94.4\\
    CLIP-Vit+Focal Loss & 92.7 &  89.7 &  \textbf{93.8} & 72.2&95.3\\
    ConvNeXt+Linear Probe & 93.7 & 90.6  & 89.2 &  75.9 & 96.2\\ 
    ConvNeXt+Focal Loss & \textbf{94.3} & \textbf{92.4} & 91.3 & \textbf{75.8} & \textbf{97.7} \\
    \hline
\end{tabular}
\label{tab:classifier-acc}
\vspace{-2mm}
\end{table}

\begin{table*}[h]
    \centering
    \caption{Results on our benchmark. \textbf{Bold} and \underline{underlined} numbers indicate the highest and second-highest metric values, respectively.}
    \vspace{-2mm}
    \resizebox{0.8\textwidth}{!}{
    \begin{tabular}{c|ccccccccc}         
    \hline
    Models & \begin{tabular}[c]{@{}c@{}}Data\\ Retrieval\end{tabular} & \begin{tabular}[c]{@{}c@{}}Find \\  Extremum \end{tabular} & \begin{tabular}[c]{@{}c@{}}Determine \\  Range \end{tabular} & \begin{tabular}[c]{@{}c@{}}Characterize \\  Distribution \end{tabular} & \begin{tabular}[c]{@{}c@{}}Find \\  Anomalies \end{tabular} & \begin{tabular}[c]{@{}c@{}}Find \\  Clusters \end{tabular} & \begin{tabular}[c]{@{}c@{}}Find \\  Correlations/ \\  Trends \end{tabular} & \begin{tabular}[c]{@{}c@{}}Make\\ Comparisons \end{tabular}  &                                                                                      \\ \hline
    LLaVA1.6-34b & 37.69 & 35.83& 3.85  & 20.00 & 21.43 & 27.27 &51.95 & 48.84\\ 
    GPT-4-vision-preview   & \textbf{56.92} & \textbf{60.96} & \underline{30.77} & \textbf{36.67} & \textbf{42.86} & \textbf{36.36} & \textbf{68.83} & \underline{56.40}\\
    Qwen-VL-Plus    & 43.08 & 21.39 & 11.54 & 10.00 & 7.14 & 13.64 & 41.56 & 34.30 \\
    Our model  & \underline{46.15} & \underline{53.48} & \textbf{35.57} & \underline{30.00} & \textbf{42.86} & \textbf{36.36} & \underline{64.94} &  \textbf{58.14}\\
    \hline
    \end{tabular}
    }
    \vspace{-3mm}
    \label{tab:our-benchmark-result}
\end{table*}

\begin{table}[t]
    \centering
    \caption{Results on traditional benchmarks (\ie, ChartQA and Chart-to-table). We compare our work with the previous open-source models and present results of ablations on data, training, and model design.}
    \vspace{-2mm}
\begin{tabular}{c|cc c |c}
    \hline
     & & ChartQA & & \\ 
     Model  & Aug. & Human & Average & Chart-to-table\\ 
    \hline
    Chart-T5 & 74.4 & 31.8 & 52.95  & 37.5\\
    Donut & 78.1 & 29.8 & 53.95 & 38.2\\
    Matcha & 88.9 & 38.8 & 63.85 & 39.4\\
    Unichart & 87.8 & 43.9 & 65.85 &  91.1  \\ 
    ChartLLaMa & 90.4 & 48.9 & 69.7&  90.0 \\ 
    ChartAst-D (39.4M CQA data) & 91.3 & 45.3 & 68.3 & \textbf{92.0} \\
    ChartAst-S (39.4M CQA data) & 92.0 & 58.2 & 75.1 & 91.6 \\
    \hline
    No Unfreezing vision encoder &  77.4 & 47.1  &  62.3 & 44.6\\
    No High Resolution  & 88.6 & 55.8 & 72.2 & 87.9 \\
    No Filtered Data & 91.5 & 60.9 & 76.2 & 90.9 \\
    No Generation Data   & 92.6 & 62.7 & 77.65 & 91.2\\
    Our model (199K CQA data) & \textbf{93.6} & \textbf{63.6} & \textbf{78.6} & 91.8\\
    \hline
\end{tabular}
\vspace{-3mm}
\label{tab-chartqa-results}
\end{table}

\subsection{Comparison to the State of the Art}

\subsubsection{Benchmarks}
\noindent \textbf{ChartQA.} The dataset information, the \textit{relaxed accuracy} metric, and the prompt for short answers have been illustrated in Sect.~\ref{sssec:bias_impact}.
ChartQA's training set is included in our training data.

\noindent \textbf{Chart-to-table.}
For evaluating MLLM's recognition ability towards chart, we follow the evaluation framework of DePlot\cite{liu2023deplot} and report the \emph{F1} scores of chart-to-table data extraction,
which measures the similarity of tables by comparing their structure and values but is invariant to column/row permutations.

\noindent \textbf{Our benchmark.}
For evaluating MLLM's performance across real-world charts and task distribution, we adopt the benchmark and its corresponding metric established in Sect.~\ref{ssec:dataset-benchmark}.

\subsubsection{Baselines}
We choose and organize MLLMs into two groups for our comparison experiments. 
For traditional benchmarks, we benchmark our models against traditional chart-specialized models, including Chart-T5\cite{masry2022chartqa}, Donut\cite{kim2022donut}, Matcha\cite{liu2023matcha}, Unichart\cite{masry2023unichart}, ChartLlama\cite{han2023chartllama}, and ChartAssistant\cite{meng2024chartassisstant}.
For our benchmark, which tests for tasks not typically included in previous chart-specific model training, we compare with SOTA generic models, including LLaVA1.6-34b\cite{liu2023improvedllava}, GPT-4-vision-preview\cite{gpt4-vision-preview}, and Qwen-VL-Plus\cite{bai2023qwen}.

\subsubsection{Results}
Table~\ref{tab-chartqa-results} presents the results of our model's performance against other models.
It demonstrates that our model consistently outperforms the baseline across all tasks. 
Particularly, we surpass the current leading models while utilizing significantly less data, showcasing our data filtering and generation effectiveness.
Moreover, the ablation studies present the effectiveness of \textit{unfreezing vision encoders} and  \textit{mixing vision encoders for high-resolution}.

Table~\ref{tab:our-benchmark-result} showcases comparative results on our benchmark, illustrating that our model outperforms baseline models in most tasks. 
These results highlight our model's strong performances, which are evident in frequently encountered tasks like \emph{data retrieval} and less common tasks such as \emph{determine range} and \emph{characterize distributions}. 
\revise{Nevertheless, current MLLMs, including our model, still exhibit subpar performance in specific tasks. 
For instance, \emph{find anomalies} and \emph{find clusters}, associated with scatterplots and bubble charts, remain challenging. 
Both chart types encode data using points, which necessitates extremely powerful recognition capabilities to discern and correlate data-visual mappings at a fine-grained level due to the small size of the visual "point."
This underscores the difficulty of certain CQA tasks that demand precise recognition of small graphical elements, particularly when addressing the challenge of just-noticeable-difference problems\cite{haehn2018evaluating, lu2021modeling}.
In contrast, for tasks like \emph{find correlations/trends} of scatterplots or line charts, the answer space is limited (\eg, positive correlation and negative correlation), and they can be inferred based on the overall feature of images rather than specific small areas of images.
}

\subsubsection{Cases}
Figure~\ref{fig:teaser} shows the comparison of state-of-the-art MLLMs (Qwen-VL-Max\cite{bai2023qwen} and GPT-4-vision-preview\cite{gpt4-vision-preview}) with our model on common difficult chart questions requiring a fine-grained understanding of visual encodings.
In the first example, the line chart has an inverted y-axis, which confuses the other models.
The second bar chart example contains a truncated y-axis that introduces recognition difficulty.
In the third bubble chart example, other MLLMs cannot understand the mapping between the bubble size and the number of employees. 
In the last example, GPT-4-vision-preview and Qwen-VL-Max also misunderstand the meaning of the stacked area.
In comparison, our model can successfully cope with these questions because of its enhanced understanding of visual encodings.

\section{Discussion}

\revise{
\noindent \textbf{Visual encoder enhancement}.
Our research finds that unfreezing the vision encoders substantially enhances the chart recognition capabilities of MLLMs, showing the original CLIP-Vit's under-performance in chart images\cite{xiao2023wytiwyr}.
An intuitive alternative design is replacing CLIP-Vit with an encoder pre-trained specifically on chart images.
For example, ChartInstruct-LLaMA \cite{masry2024chartinstruct} substitute the CLIP-Vit in LLaVA with the UniChart vision encoder\cite{masry2023unichart}.
However, the researchers did not observe model performance improvements compared to Unichart.
This highlights the superiority of generic vision encoders, which learn robust image interpreting capabilities (\eg, localization) from millions of natural images.
}
Moreover, understanding some real-world charts requires broad vision knowledge.
For instance, infographics may incorporate natural images to depict certain chart elements\cite{xiao2023let} vividly.
\revise{Borrowing LLaVA-Med's lesson\cite{li2024llava} in initializing CLIP-Vit, developing a visualization-domain CLIP with enhanced basic chart interpretation performance is a promising future work.}

\noindent \textbf{Better textual representation for chart understanding}.
Aligning the language model with the vision encoder is crucial, and dense image representations, such as high-quality captions, play an essential role in this process. 
Typically, data tables are used for charts due to their abundant information. 
However, the intrinsic limitation of the data table is the loss of all visual information.
While captions for charts keep certain visual information, the numerical information is hard to keep completely.
Vistext\cite{benny2023vistext} has explored the use of scene graphs as a potential alternative to data tables. 
Despite this, the choice of data format for a language model is a significant consideration, and it remains to be thoroughly investigated whether the scene graph format can effectively integrate within the context of MLLMs.

\noindent \revise{\textbf{Insights on applying MLLMs to complex reasoning visualization tasks}.
Our research finds that current MLLMs still face challenges in analytical tasks (\eg, \emph{find anomalies} and \emph{determine range}).
Recently, referential question-answering \cite{zhang2023gpt4roi, chen2023shikra}  has been shown to benefit the comprehension of complex spatial relationships. 
It requires annotating bounding boxes and arrows in images and referring to these elements in questions.
This task was not considered in our training data because of the lack of chart data with referential annotations.
However, referential QAs are common in real-world visual analytics and potentially beneficial for tasks like \emph{find anomalies}.
For example, we can use bounding boxes to label anomaly points or highlight a range of data elements, boosting MLLMs' understanding of relevant tasks.
We place the exploration of referential and other possible complex QA formats tailored for charts in future work.
Furthermore, end-to-end MLLM outputs inherently pose uncertainty.
Incorporating golden tables and code of charts with MLLM for interaction applications may be robust in complex visual analytics scenarios.
}

\section{Conclusion}
This study addresses significant challenges in advancing MLLMs' performances in CQA.
An empirical study is conducted to investigate the limitations of existing MLLMs and CQA datasets. 
We identify a critical need for fine-grained consideration of visual encodings and QA tasks, which current data collection and synthesis methods overlook, leading to unbalanced data distribution and inconsistent data quality. 
With a two-stage data engine of filtering-then-generation, we filter existing datasets and enlarge them through LLM-based generation techniques, ensuring a broader range of high-quality data that captures the characteristics of charts. 
By incorporating a mixture-of-resolution adaptation strategy and unfreezing the vision encoder during model training, we significantly improve the performance of our MLLM on CQA tasks. 
Experiments demonstrate that, even with a more compact dataset, our model surpasses SOTA CQA models, highlighting the efficacy of our methodology. 
We also contribute a benchmark for future advancements in MLLMs for CQA tasks.

\acknowledgments{
This work is supported partially by National Natural Science Foundation of China (62172398), and the Guangzhou Basic and Applied Basic Research Foundation (2024A04J6462, 2023A03J0142).
}


\bibliographystyle{abbrv-doi-hyperref}

\bibliography{references}

\end{document}